\documentclass[sn-nature,Numbered]{sn-jnl}

\usepackage{soul}
\usepackage[table]{xcolor}%
\usepackage[most]{tcolorbox}

\usepackage{multirow}
\usepackage{url}
\usepackage[utf8]{inputenc}
\usepackage{graphicx}%
\usepackage{multirow}%
\usepackage{amsmath,amssymb,amsfonts}%
\usepackage{amsthm}%
\usepackage{mathrsfs}%
\usepackage[title]{appendix}%

\usepackage{textcomp}%
\usepackage{manyfoot}%
\usepackage{booktabs}%
\usepackage{algpseudocode}%
\usepackage{listings}%
\usepackage[linesnumbered,ruled,vlined]{algorithm2e}

\theoremstyle{thmstyleone}%
\theoremstyle{thmstyletwo}%
\theoremstyle{thmstylethree}%
\begin{document}

\title[Forecasting Intraday Power Output by a Set of PV Systems using Recurrent Neural Networks and Physical Covariates]{Forecasting Intraday Power Output by a Set of PV Systems using Recurrent Neural Networks and Physical Covariates}

\author*{\fnm{Pierrick} \sur{Bruneau}*}\email{pierrick.bruneau@list.lu}
\author{\fnm{David} \sur{Fiorelli}}\email{david.fiorelli@list.lu}
\author{\fnm{Christian} \sur{Braun}}\email{christian.braun@list.lu}
\author{\fnm{Daniel} \sur{Koster}}\email{daniel.koster@list.lu}

\affil{\orgname{Luxembourg Institute of Science and Technology}, \orgaddress{\street{5 Avenue des Hauts-Fourneaux}, \city{Esch-sur-Alzette}, \postcode{L-4362}, \country{Luxembourg}}}

\abstract{
Accurate intraday forecasts of the power output by PhotoVoltaic (PV) systems are critical to improve the operation of energy distribution grids. We describe a neural autoregressive model that aims to perform such intraday forecasts.
We build upon a physical, deterministic PV performance model, the output of which is used as covariates in the context of the neural model. In addition, our application data relates to a geographically distributed set of PV systems. We address all PV sites with a single neural model, which embeds the information about the PV site in specific covariates. We use a scale-free approach which relies on the explicit modeling of seasonal effects.
Our proposal repurposes a model initially used in the retail sector and discloses a novel truncated Gaussian output distribution. 
An ablation study and a comparison to alternative architectures from the literature shows that the components in the best performing proposed model variant work synergistically to reach a skill score of 15.72\% with respect to the physical model, used as a baseline.
}


\maketitle

\section{Introduction} \label{sec:intro}

Grids of PV systems have become an inevitable component in the modern and future energy distribution systems. However, weather conditions cause the magnitude of PV power production to fluctuate and consumer supply has to be adapted to demand at each point in time.
Distribution system operators (DSOs) have increasing and specific requirements for PV power forecasts. Indeed, fluctuating renewables could cause operational issues (e.g., grid congestion), that call for active grid operation. 
In this context, the fine-grained forecasts of PV power expected in the near future are critical in view of facilitating operations. Also, many forecasting models issue point forecasts, but barely characterize the uncertainty attached to their forecasts, where such information can be critical for a DSO in order to quantify and mitigate risks in the context of a trading strategy.

PV power production can be forecasted using a deterministic PV performance model \cite{koster_short-term_2019}. This models the internal physical mechanisms of a PV system, from the solar irradiance received to the electrical power produced using a set of physical equations (referred to as a \emph{physical model} in the remainder of this paper). Therefore, for each time step for which PV power has to be predicted, a solar irradiance forecast, provided by a Numerical Weather Prediction (NWP) service, is required as input. The underlying hypothesis is that solar irradiance is fairly smooth over limited regional areas, and the production curve specific to a PV system will be mainly influenced by how it converts this solar energy to PV power according to its specifications.
In \cite{koster_single-site_nodate}, the authors of the present paper introduced a model that performs intraday probabilistic forecasts of PV power production. It combines the abovementioned physical model with a model based on Long-Short-Term Memory (LSTM) cells \cite{hochreiter97}. This kind of combination of a model based on a set of physical equations and a statistical model is sometimes referred to as \emph{hybrid-physical} in the solar energy literature \cite{antonanzas2016}. For training and evaluation, it uses real data provided by Electris, a DSO in Luxembourg. Results show that this new model improves baseline performance, while managing local effects such as PV system shading. Our earlier paper rather targeted solar energy specialists, with few details revealed about how the machine learning model, acting as cornerstone to the approach, was designed and trained. The present paper aims to fill this gap by providing entirely new material that focuses on this complementary view. Specifically, the purpose of the present paper is to focus on neural time series forecasting aspects in the context of this application.

The primary contributions of the present work focus on designing a model architecture and a training procedure that meet the operational needs expressed by the DSO. In Section \ref{sec:related}, we provide a structured survey of the related work in order to position the applicative problem addressed, and determine which existing work could be reused or repurposed to suit our needs. We identify the DeepAR model \cite{salinas_deepar_2020} as the most suitable for our case study, so for self-consistency, we introduce the required formalism and summarize its inference and loss logic in Section \ref{sec:deepar}.

The contributions of our applicative paper are:
\begin{itemize}
    \item Proposing a novel truncated Gaussian output component, which we integrate in DeepAR, as detailed in Section \ref{sec:positivegaussian},
    \item Casting the hybrid-physical approach in terms of statistical model covariates as described in Section \ref{sec:data},
    %\item Using PV system specifications as covariates in a univariate model,
    \item Carefully examining the implications of our case study on the computation of model loss and the relevance of the validation scheme, as discussed in Sections \ref{sec:loss} and \ref{sec:validation}, respectively.
\end{itemize}

We provide a thorough experimental evaluation in Section \ref{sec:experiments}. Several variants of our model are compared to alternative models from the literature, and an ablation study allows us to emphasize the specific contribution of each of its components. Finally, in Section \ref{sec:discussion}, we highlight qualitative results to demonstrate how our approach mitigates local effects that impact the PV performance model.

\section{Related Work} \label{sec:related}

In Section \ref{sec:pvpower}, we review seminal PV power forecasting methods. Then in Section \ref{sec:ml}, we survey the time series forecasting addressed in the machine learning (ML) literature from the perspective of its repurposing potential for the PV power forecasting application. Finally, Section \ref{sec:validation} focuses on the distinctive features in terms of forecasting structure and validation that come with ML approaches applied to time series forecasting.

\subsection{PV Power Forecasting} \label{sec:pvpower}

Most approaches in PV power forecasting model the conversion chain of solar irradiance to electrical power in a PV system and therefore follow a two-step approach: first, they forecast the solar irradiance, then they convert it to PV power forecasts \cite{blaga2019, koster_short-term_2019}. 
The most common way of forecasting solar irradiance relies on NWP systems such as the European Centre for Medium-Range Weather Forecasts (ECWMF) Ensemble Prediction System \cite{molteni_ecmwf_1996}. It issues hourly regional forecasts for a range of meteorological variables (including solar irradiance) for the 10 days to come, thus including the intraday range. 
Intraday is sometimes defined in the literature as a forecast horizon of up to 6 hours \cite{antonanzas2016}. In this paper, for experimental simplicity, we deviate from this definition by considering intraday as the 24 hours to come, starting at midnight of the same day.
The PV power data described in Section \ref{sec:data} is collected each day at 06:00 CET by Electris. As we exclude night-time slots from the metrics presented in Section \ref{sec:loss}, our intraday definition is roughly equivalent to that used in the context of the Nord Pool power exchange market\footnote{\url{https://www.nordpoolgroup.com/}}.
In view of improving the NWP system outputs, or to avoid having to rely on such systems, solar irradiance can also be forecasted using statistical methods and ML models. The simplest of these include persistence models, which are often adjusted using clear sky models \cite{yang_choice_2020}. \cite{inman_solar_2013} also review various ML techniques that have been employed for this purpose, e.g., AutoRegressive (AR) models, Feed-Forward Networks (FFN) and k-Nearest Neighbors. In this range of contributions, \cite{bruneau_bayesian_2012} address the intraday hourly prediction of solar irradiance using an ensemble of FFN models. Specifically, they implement rolling forecasts by specializing each model on a current time and a prediction horizon.

PV power forecasting is reviewed in detail by \cite{antonanzas2016}. In this domain, several approaches aim to model the series of PV power values directly, without having to rely on solar irradiance forecasts. \cite{elsinga_short-term_2017} propose short-term forecasts ($<$30mn), which exploit the cross-correlation of PV measurements in a grid of PV systems. They hypothesize that clouds casting over a given PV system have a lagged influence on other systems downwind. They optimize the associated time lag and cloud motion vector.
\cite{lonij_intra-hour_2013} also consider a spatially distributed set of PV panels. They directly use PV power values, without converting proxy information such as solar irradiance or cloud density. Similarly to \cite{elsinga_short-term_2017}, they focus on correlations among stations to help account for intermittency due to clouds.

\cite{pedro_assessment_2012} present AR approaches to PV power forecasting. They focus on forecasting one and two hours ahead, where NWP models tend to under-perform. Several models are compared, including persistence, linear models such as AutoRegressive Integrated Moving Average (ARIMA) \cite{hyndman_automatic_2008}, and FFN. 
They found out that FFN perform the best, with improvements provided by the optimization of FFN parameters, input selection and structure using a Genetic Algorithm. 
%They conjecture that binning data according to the associated season, and learning per-bin models should improve forecasting ability overall, even if this approach has been recently criticized \cite{salinas_deepar_2020}.

\subsection{ML approaches for time series forecasting} \label{sec:ml}

Among other related work, Section \ref{sec:pvpower} surveyed contributions that used ML methods in view of forecasting solar irradiance and PV power production. In this section, we generalize this view by surveying recent work in time series forecasting at large. The methods in this section were generally not applied to the applicative context considered in the present paper, but could be repurposed \emph{a priori}.
Besides neural and ARIMA models, seminal ways to forecast time series include the Croston method, which is an exponential smoothing model dedicated to intermittent demand forecasting \cite{shenstone_stochastic_2005}. It is notably used as a baseline method in \cite{salinas_deepar_2020}, along with the Innovation State-Space Model (ISSM) \cite{seeger_bayesian_2016}.

Modern, so-called \emph{deep} neural network architectures, such as the Long-Short-Term Memory (LSTM) \cite{hochreiter97} and the Gated Recurrent Unit (GRU) \cite{cho14}, exploit the sequential structure of data.
Although recurrent models such as the LSTM would appear outdated, they are still popular in the recent literature thanks to improvements in training and inference procedures pushed by modern toolboxes such as Tensorflow \cite{tensorflow2015-whitepaper} or MXNet \cite{chen15}, as well as the encoder-decoder mechanism, in which a context sequence is encoded and conditions the prediction of the target sequence. It was initially codified for the GRU, and then transferred to the LSTM, leading to its continued usage in recent contributions \cite{shi_convolutional_2015, salinas_deepar_2020}.
These models contrast with the seminal Feed-Forward Network (FFN), in which all layers are fully connected to the previous and next layers, up to the activation layer \cite{bebis_feed-forward_1994}.

Salinas et al. propose DeepAR, which implements flexible forecasting for univariate time series \cite{salinas_deepar_2020}. Formally, it defines a \emph{context} interval (a chunk of past values) and a \emph{forecasting} interval (the set of values to be forecasted), and the model is optimized end-to-end with regards to the whole forecasting range. This contrasts with linear models such as ARIMA, which are optimized for one time step ahead forecasts.
Also, instead of point forecasts, DeepAR predicts model parameters, which can be used to compute sample paths and empirical quantiles, and are then potentially used by the DSO to adjust its trading strategy in the context of this paper. In this case, a family of probability distributions has to be chosen so as to fit the time series at hand. The model was initially aimed at retail business applications, but can be adapted to other types of data just by changing the family of the output probability distribution. It is based on the LSTM model architecture. 
The model supports the adjunction of covariates, i.e., time series which are available for both context and forecasting intervals at inference time. By repeating the same value for all intervals, static covariates are also supported.
%Such flexibility meets all the requirements of our hybrid-physical approach (i.e., NWP covariates and PV system descriptive features).

In retail applications, input data may have a highly variable magnitude (e.g., depending on item popularity or time of year). The authors observe an approximate power law between item magnitude and frequency (i.e., values are less likely as they get bigger, and vice versa). They claim that grouping items to learn group-specific models or performing group-specific normalizations, as previously done in the solar and PV power forecasting literature \cite{pedro_assessment_2012,bruneau_bayesian_2012} are not good strategies in such cases and propose a simple alternative scheme, where samples are scaled by a factor computed using the context interval.

%\cite{fawaz_inceptiontime_2019} propose a time series classification model inspired by Inception-v4 \cite{szegedy_inception-v4_2016}. Basically, they are  transferring the multi-scale pattern extraction capability of convolutional neural networks to 1D data such as time series. 
A convolutional encoder was tested by Wen et al. in the context of their multi-horizon quantile forecaster \cite{wen_multi-horizon_2018}. Instead of forecasting probabilistic model parameters, this model directly forecasts quantiles in a non-parametric fashion. However, it suffers from the quantile crossing problem: forecasted values may have ranks inconsistent with the quantile they are attached too. 
\cite{gasthaus_probabilistic_2019} is another alternative to DeepAR. Similarly to \cite{wen_multi-horizon_2018}, it does not rely on probability distribution outputs, and rather implements conditional quantile functions using regression splines. Spline parameters are fit using a neural network that directly minimizes the Continuous Ranked Probability Score (CRPS) \cite{gneiting_calibrated_2005}, which is then used as a loss function. This results in a more flexible output distribution, and is an alternative to other flexible schemes (e.g., mixture of distributions in the context of \cite{salinas_deepar_2020}). However, it currently\footnote{Checked on 21 June 2023} lacks a publicly available implementation.

Multivariate forecasting consists of modelling and forecasting multiple time series simultaneously, in contrast to univariate forecasting. The seminal way to achieve this, is with the Vector AutoRegression (VAR) model, which is an extension of the linear autoregressive model to multiple variables \cite{stock2016}. As this model finds it difficult to deal with many variables (e.g., items in the retail domain), neural network-based models such as DeepVAR were designed as an alternative \cite{salinas_high-dimensional_2019}. It can be thought of as a multivariate extension to \cite{salinas_deepar_2020}.
DeepVAR models interaction between time series, e.g., as it results from causality or combined effects. It uses a Copula model, which models interactions between time series, and elegantly copes with time series of varying magnitude, alleviating the need for an explicit scaling mechanism. A single multivariate state variable underlying a LSTM model is used for all time series. 
The authors also define a low-rank parametrization, which opens the possibility of dealing with a very large number of time series.

Some prior work involved deep learning in the context of PV power forecasting. For example, \cite{abdel-nasser_accurate_2019} consider one hour ahead forecasts (instead of intraday as it is the aim in this paper). They used a single LSTM layer without the encoder-decoder mechanism, limited to point forecasts. Data for two PV sites is used for the experiments, with roughly the same power magnitude for both sites (approx. 3.5kW). Models are trained for each site separately. Alternatively, in this paper we address all sites with a single model in a scale-free approach, dealing with an arbitrary number of sites with little to no model size overhead. Finally, the locations associated with the datasets are subject to a dry climate, which is simpler to predict \cite{pedro_assessment_2012}. Our application testbed is a temperate area, subject to frequent and abrupt changes on a daily basis, and is therefore much more challenging to predict.
\cite{wang_day-ahead_2020} also address PV power forecasting, by decorrelating scale-free forecasts using LSTM, from seasonal effects modeled separately using time correlation features and partial daily pattern prediction. However, they focus on forecasts aggregated at a daily scale, while in this paper, we consider hourly data. In addition, our approach is end-to-end, with seasonal effects modeled as time covariates, as allowed by the DeepAR model \cite{salinas_deepar_2020}.

%\subsection{Hybrid-physical approaches} \label{sec:hybrid-physical}

%\cite{barbounis06} focus on wind speed and power forecasting. They consider hourly forecasts for 72 hours ahead, using Numerical Weather Prediction (NWP) forecasts as additional inputs, thus proposing an early combination of observation data with NWP covariates. The recurrent model used then (diagonal recurrent neural networks \cite{ku95}) was superseded by alternative models such as LSTM \cite{hochreiter97} and GRU \cite{cho14} in the recent literature as seen in the previous section. 

\cite{thaker_comparative_2022} present a regional PV power forecasting system, with hourly resolution up to 72 hours ahead. Their approach combines clustering and numerical optimization, and is compared to regression methods such as ElasticNet \cite{zou2005}, SARIMAX \cite{suhartono2017}, or Random Forests \cite{suhartono2017}.
Their approach is not autoregressive, rather they directly predict future PV power from solar irradiance and temperature forecasts obtained from a proprietary system which refines NWP forecasts according to local conditions. Alternatively, our approach tries to combine the benefits of using physical model forecasts (fed periodically by a NWP system) as covariates in an autoregressive model of the PV power time series. We observe that this effectively implements the \emph{hybrid-physical} approach, which combines the output of a physical model with a machine learning model \cite{antonanzas2016}. In an early work, Cao and Lin \cite{cao_study_2008} used NWP covariates in the context of a neural network to perform next hour and aggregated daily solar irradiance forecasting. More recently, several models have allowed continuous covariates to be combined with each time slot in a forecasting interval \cite{wen_multi-horizon_2018,salinas_deepar_2020,rasul_autoregressive_2021,park_learning_2022}. The model described in Section \ref{sec:model}, and the experiments in Section \ref{sec:experiments}, involve models from this family.

\subsection{Forecast structure and validation} \label{sec:validation}

\begin{figure}[h]
\centering
\includegraphics[width=.9\linewidth]{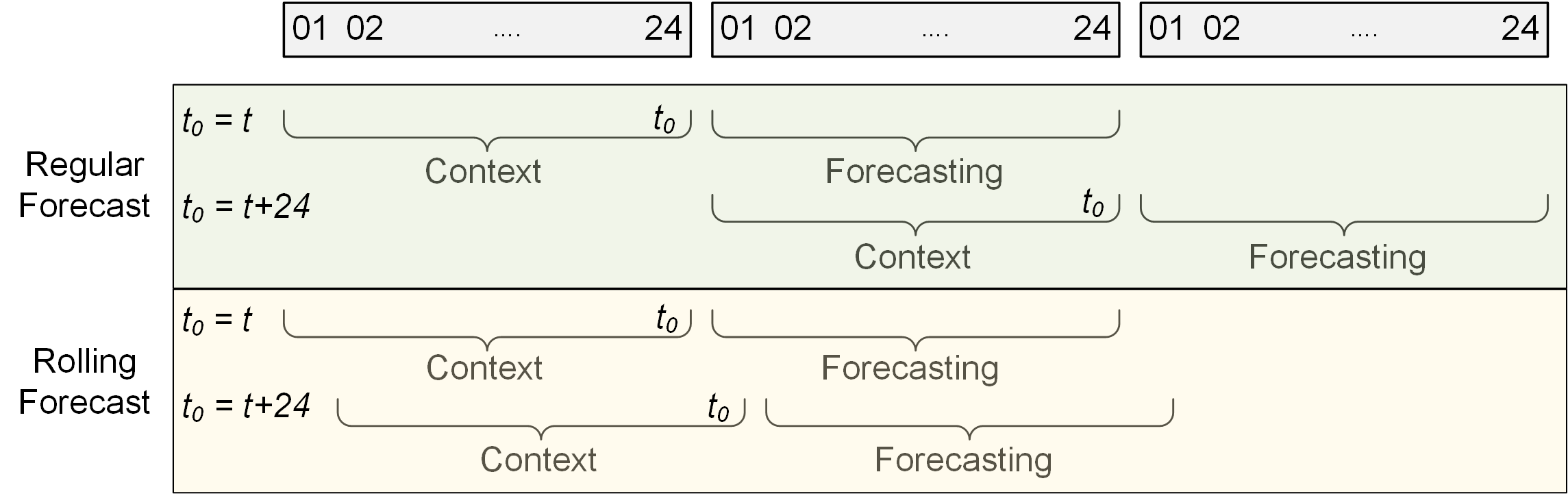}
\caption[Caption]{Distinction between regular and rolling forecasts. $t_0$ denotes the present time step in the context of a given data sample.}
\label{fig:rolling}
\end{figure}

Figure \ref{fig:rolling} distinguishes \emph{regular} forecasts from \emph{rolling} forecasts, which are the two main strategies to consider when extracting fixed-sized blocks from time series \cite{benidis_deep_2023}. For simplicity, the figure considers hourly data and a 24-hour context and forecasting intervals.
In brief, two consecutive regular forecasts are offset by the size of the forecasting interval, while rolling forecasts are offset by the frequency of the time series (hourly, in Figure \ref{fig:rolling}). In other words, with 24-hour forecasting and context intervals, regular forecasts happen at the same time every day, whereas rolling forecasts are issued every hour for the whole forecasting interval. In this process, forecasts beyond the next hour are refreshed every hour.

Works such as \cite{thaker_comparative_2022} consider regular forecasts, as the forecast time is dependent on the availability of NWP data. 
As the physical covariates we use in our work are also dependent on forecasts issued by a NWP service, and the data we use in our experiments in Section \ref{sec:experiments} is only collected daily by the DSO, regular forecasts are also a requirement in our work. Alternatively, \cite{bruneau_bayesian_2012} address rolling forecasts by having a distinct model for each possible starting time in the day. 
Let us note that some models (e.g., \cite{salinas_deepar_2020}) allow seasonal information on the predicted time steps to be encoded (e.g. hour in day, day in week) as covariates (referred to as \emph{time covariates} later on). Therefore, they can be used indistinctively with regular and rolling forecasts, provided that an adapted training set is available. Also, contrasting to \cite{wang_day-ahead_2020}, this means that modelling periodicity and seasonality explicitly is not needed, as the model then directly combines this additional input to time series data.

\cite{bergmeir_use_2012} discuss the problem of cross-validation, and more generally validation, in the context of time series forecasting. Original formulations of cross-validation methods often assume that data items are independent. They cannot be used out of the box with time series, as the sequential structure of the latter invalidates the underlying hypotheses. 
They recognize that the seminal way to validate models with time series is to train a model using the first $t_0$ values, and validate using the last $T-t_0$ values. However, this method is hardly compatible with cross-validation schemes, and yields weak test error estimations. 
In virtue of the bias-variance tradeoff \cite{geman_neural_1992}, this issue has a moderate impact on models with a strong bias, such as linear models. However, over-parametrized models, such as most neural models presented in Section \ref{sec:ml} (e.g. \cite{salinas_deepar_2020,salinas_high-dimensional_2019,gasthaus_probabilistic_2019,wen_multi-horizon_2018}) can be significantly affected, and exhibit a strong tendency to overfit. 
For mitigation, \cite{bergmeir_use_2012} recommend blocked cross-validation, in which time series segments of size $\tau << T$ are used for independent training and validation for model selection and test error computation. As we also use deep learning as a building block in our approach, we carefully consider these recommendations in our experimental design (see Section \ref{sec:experiments}).

\section{Model Description} \label{sec:model}

The survey of related works in Section \ref{sec:related} led us to choose DeepAR \cite{salinas_deepar_2020} as a framework to develop our implementation. We adapted the official implementation of the model \cite{gluonts_jmlr} to suit our needs. Another model which offers the relevant flexibility as well as public implementation is that of Wen et al. \cite{wen_multi-horizon_2018}. We will make a comparison with this model in our experimental section.

Alternatively, PV sites could have been considered as dimensions in a multivariate forecasting problem, thus possibly forecasting all sites at a given time at once using DeepVAR \cite{salinas_high-dimensional_2019}. However, its limitation to static covariates precludes the implementation of our proposed approach, which relies on using dynamic forecasts from the physical model as covariates to enhance predictive capabilities.

Also, it is unclear how new PV sites and missing values, which are quite common due to independent breakdowns and interruptions of measurements or data communication at PV sites (see Section \ref{sec:scheme} for a quantitative account), can be handled with such multivariate modelling. Therefore, we chose to model the PV site using covariates in the context of a univariate model. We will test the value of these covariates in our ablation study in Section \ref{sec:experiments}.

\subsection{DeepAR model} \label{sec:deepar}

In the remainder of the paper, for clarity purposes, scalar variables are represented in normal font, vector variables in bold font, and matrix variables in capital bold font. Section \ref{sec:deepar} essentially paraphrases \cite{salinas_deepar_2020}, but ensures the present paper is self-contained while introducing the necessary formalism in general terms.

Let us assume we have a data set of $N$ univariate time series, each with a fixed size $T$. Each observed time series in $\{ \bold z_n \}_{n \in 1, \dots, N}$ may relate to an item in store in the retail context, or to a distinct PV system in the context addressed in this paper. 
$t_0 \in [1, \dots, T]$ denotes the present time, i.e. the latest time point for which we assume $z_{n,t}$ is known when issuing the forecast. $[1, \dots, t_0]$ is then the \emph{context} interval, and $[t_0+1, \dots, T]$ is the \emph{forecasting} interval. 
Let us note that each $t \in [1, \dots, T]$ is merely associated with the position of values within each sample $\bold z_n$; this is not inconsistent with the fact that $t$, in the context of two different samples $n$ and $n'$, may be linked to different \emph{absolute} points in time, e.g., referring to two different days if grouping the data together as described in Section \ref{sec:scheme}.
The goal of the model is then to forecast $\bold z_{n, t_0+1:T} = [z_{n,t_0+1}, \dots, z_{n,T}]$ with the knowledge of $\bold z_{n, 0:t_0} = [z_{n,0}, \dots, z_{n,t_0}]$. We also consider a set of covariates $\bold X_{n,1:T} = [\bold x_{n,1}, \dots, \bold x_{n,T}]$ which are known for $t \in [1, \dots, T]$, even at time $t_0$.

\begin{figure}[h]
\centering
\includegraphics[width=.9\linewidth]{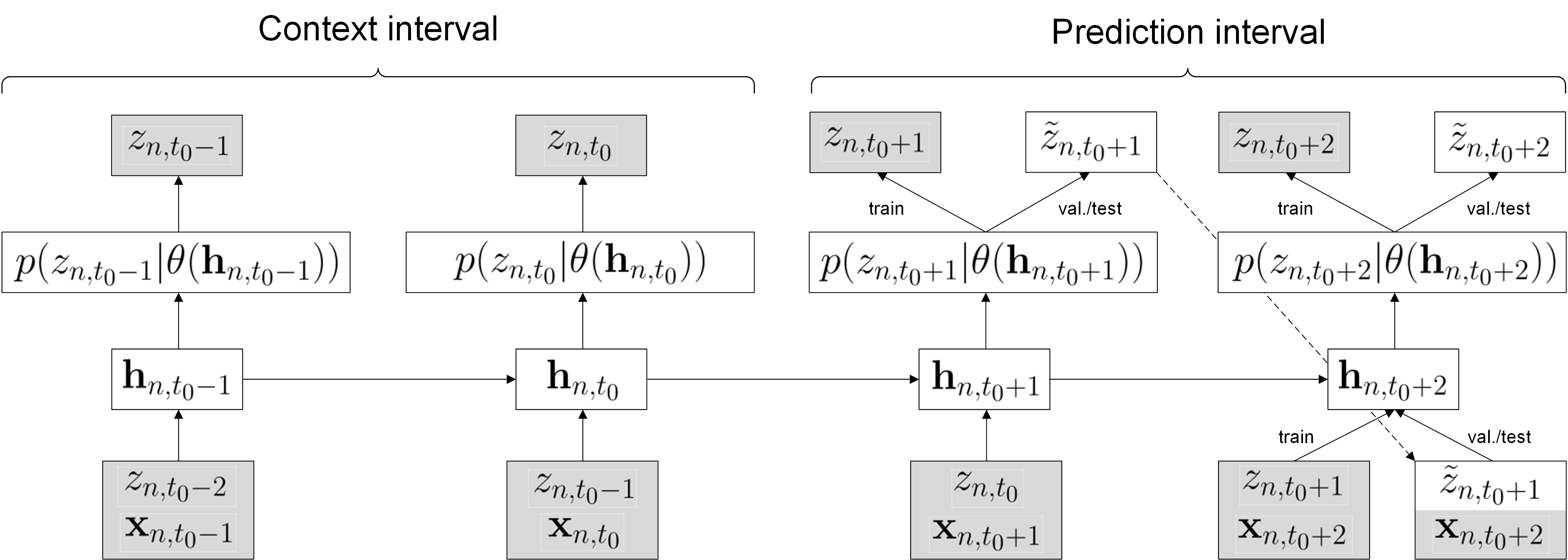}
\caption[Caption]{Illustration of the DeepAR model. Observed variables are represented as shaded boxes, and latent variables as blank boxes. For the context interval, $z$ variables are always known. For the forecasting interval, the model behaves differently at training and test time. At test time, $\tilde z$ variables are sampled according to $p$, forming sample paths. Plain lines represent dependencies between random variables, and the dashed line highlights the reinjected sample.}
\label{fig:deepar}
\end{figure}

In this context, the model is defined by the following product of likelihood factors, also summarized in Figure \ref{fig:deepar}:

\begin{align}
    Q_\Theta &= \prod_{n=1}^N \prod_{t=t_0+1}^T {q_\Theta(z_{n,t} \vert \bold z_{n,1:t-1}, \bold X_{n,1:T})} \nonumber \\
     &= \prod_{n=1}^N \prod_{t=t_0+1}^T { p(z_{n,t} \vert \theta(\bold h_{n,t}, \Theta)) }
    \label{eq:likelihood_model}
\end{align}

The model is both autoregressive and recurrent, as the state variable:

\begin{equation}
    \bold h_{n,t} = \Theta(\bold h_{n,t-1}, z_{n,t-1}, \bold x_{n,t}) \label{eq:recursion}
\end{equation}

is obtained from LSTM model $\Theta$ in which the state variable and observation of the previous time step are both reinjected. The model also depends on the parametrized function $\theta$, which learns the mapping between the state variable $\bold h$ and parameters of the probability distribution $p$. 

In effect, as seen in Figure \ref{fig:deepar}, when training the model, $\bold z_n$ is known for all $t \in [1 \dots T]$, even the forecasting interval. 
The $z_{n,t}$ boxes at the top of the diagram serve to compute losses that are then backpropagated to the model, even in the forecasting interval (i.e., \emph{teacher forcing}).
However, at inference time, actual observations are not available for the forecasting interval, so we sample $\tilde z_{n,t} \sim p$, and inject them as proxy observations. Doing this yields sample paths, which can be repeated and serve to compute empirical quantiles in the forecasting interval, instead of simple point estimates. In this paper, when the point estimates $\hat z_{n,t}$ are needed, we take them as the empirical median of a set of sample paths. In Figure \ref{fig:deepar}, we can see that the LSTM \emph{encodes} the context interval into $\bold h$, which is then \emph{decoded} for the forecasting interval. The same LSTM model $\Theta$ is used for encoding and decoding.

The negative log of expression \eqref{eq:likelihood_model} is used as the loss function for training all parameters in the model in an end-to-end fashion. The form of function $\theta$ depends on the probabilistic model in expression \eqref{eq:likelihood_model}: for example, if $p$ is chosen as a Gaussian, the appropriate functions would be:

\begin{align}
    \theta_\mu(\bold h_{n,t}) &= \bold w_\mu \bold h_{n,t} + b_\mu \nonumber \\
    \theta_\sigma(\bold h_{n,t}) &= \log(1+\exp(\bold w_\sigma \bold h_{n,t} + b_\sigma)) \label{eq:sigmamap} 
\end{align}

We note that the softplus function in \eqref{eq:sigmamap} ensures that $\sigma$ is mapped as a positive real number. Among possible probabilistic models and mapping functions, the official DeepAR implementation \cite{gluonts_jmlr}\footnote{\url{https://github.com/awslabs/gluonts}}, used in the experiments for this paper, features Gaussian, Student, negative binomial, and mixture distributions. The mixture distribution composes several distributions from the same nature using mixture weights, which have their dedicated $\theta$ function.

Algorithms \ref{alg:training} and \ref{alg:inference} are the pseudo-codes describing the articulation of equations above in the context of a training loop and at inference time, respectively. For simplicity, we considered only one sample in the algorithm: with a data set, samples are grouped in batches and the algorithm remains the same otherwise.
At training time, the point is to compute losses for the whole interval from $1$ to $T$, while at inference time, the goal is to estimate $\tilde z_{n,t}$ for the forecasting interval.

\begin{algorithm}[H]
\SetAlgoLined
\KwResult{$\theta$ (mapping function) and $\Theta$ (LSTM)}
\KwIn{$\bold X_{n,1:T}, \bold z_{n,0:T}$}
%\KwOut{Output}
\tcc{Teacher forcing mode: $\bold z_{n,0:T}$ is completely known}
Randomly initialize $\bold h_{n,0}$\;
loss $\leftarrow$ array of size $T$\;
\ForEach{$t \in 1:T$}{
	Compute $\bold h_{n,t} \leftarrow \Theta(\bold h_{n,t-1}, z_{n,t-1}, \bold x_{n,t})$ using Eqn. \eqref{eq:recursion}\;
	Compute probabilistic parameters $\theta(\bold h_n,t)$ using Eqn. \eqref{eq:sigmamap}\;
	Compute $\text{loss}[t]$ as the negative log of Eqn. \eqref{eq:likelihood_model} with regards to $z_{n,t} \text{ and } \theta$\;
}
Back-propagate through time loss to update $\theta$ and $\Theta$\;

\caption{Pseudo-code of the model used at training time}
\label{alg:training}
\end{algorithm}

\begin{algorithm}[H]
\SetAlgoLined
\KwResult{$\tilde {\bold z}_{n,(t_0+1):T}$}
\KwIn{$\bold X_{n,1:T}, \bold z_{n,0:t_0}, n_\text{samples}$}
%\KwOut{Output}
\tcc{First, we unroll the LSTM for the context interval}
\tcc{The goal is to encode the context in $\bold h_{n,t_0}$}
Randomly initialize $\bold h_{n,0}$\;
\ForEach{$t \in 1:t_0$}{
	Compute $\bold h_{n,t} \leftarrow \Theta(\bold h_{n,t-1}, z_{n,t-1}, \bold x_{n,t})$ using Eqn. \eqref{eq:recursion}\;
}
sample\_paths $\leftarrow$ matrix of size ($n_\text{samples}, T-t_0+1$)\;
\ForEach{$i \in 1:n_\text{samples}$}{
    sample\_paths[i,1] $\leftarrow z_{n,t_0}$\;
    \ForEach{$t \in (t_0+1):T$}{
		Compute $\bold h_{n,t} \leftarrow \Theta(\bold h_{t-1}, \text{sample\_paths}[i,t-t_0], \bold x_{n,t})$ using Eqn. \eqref{eq:recursion}\;
		Compute probabilistic parameters $\theta(\bold h_{n,t})$ using Eqn. \eqref{eq:sigmamap}\;
		Sample $\text{sample\_paths}[i,t-t_0+1]$ using Eqn. \eqref{eq:likelihood_model} with $\theta$\;
	}
}
$\tilde z_{n,(t_0+1):T} \leftarrow \text{median}(\text{sample\_paths}[:,2:(T-t_0+1)])$\;

\caption{Pseudo-code of the model used at inference time}
\label{alg:inference}
\end{algorithm}

\subsection{Positive Gaussian likelihood model} \label{sec:positivegaussian}

As PV power measurements are bound to be non-negative real numbers, a contribution of this paper is to allow the Gaussian distribution to be truncated from below at 0 (the upper limit remaining $+\infty$), referred to as the \emph{positive Gaussian} distribution in the remainder of this paper. Formally this yields:

\begin{equation}
    p(z_{n,t} \vert \theta_\mu, \theta_\sigma) = \frac 1 {\sigma \sqrt{2\pi}} \frac {\exp{(-\frac 1 2 \frac{(z_{n,t} - \theta_\mu)^2}{{\theta_\sigma}^2})}}{1 - \Phi(-\frac {\theta_\mu} {\theta_\sigma})}
\end{equation}

where $\Phi$ is the cumulative distribution function of the standard Gaussian (i.e., with mean 0 and standard deviation 1). Besides adapting the loss function (see Equation \eqref{eq:likelihood_model}) to this new probability distribution function, the same $\theta_\sigma$ function as the Gaussian distribution can be used. To make sure the range of $\theta_\mu$ is also positive, for the positive Gaussian we use:

\begin{equation}
    \theta_\mu(\bold h_{n,t}) = \log(1+\exp(\bold w_\mu \bold h_{n,t} + b_\mu)) \nonumber
\end{equation}

From an application of the Smirnov transformation \cite{devroye_nonuniform_2006} to the case at hand, samples from a positive Gaussian distribution can be obtained as:

\begin{equation}
    \tilde z = \Phi^{-1} \Biggl(\Phi \bigl(-\frac {\theta_\mu} {\theta_\sigma} \bigr) + \tilde u \biggl (1 - \Phi \bigl (-\frac {\theta_\mu} {\theta_\sigma} \bigr) \biggr )\Biggr) \theta_\sigma + \theta_\mu
\end{equation}

where $\tilde u$ is a uniform sample in $[0,1]$.

\section{Experiments} \label{sec:experiments}

\subsection{Data} \label{sec:data}

Section \ref{sec:model} presented the forecasting model underlying our experiments in general terms, but here we recall that we focus on a specific application and its distinctive features: forecasting the power output of a set of PV systems.

The variable to forecast ($z$ in Section \ref{sec:model}) is the average power output of a PV system during the hour to come in Watts. As hypothesized in Section \ref{sec:validation}, it is thus an hourly time series. For our experiments, we used data recorded by 119 PV systems located in Luxembourg between 01/01/2020 and 31/12/2021. They were dispatched in a relatively small (4 $\times$ 4 km) area. These PV systems are managed by Electris, a DSO in Luxembourg which collaborated with the authors of this paper in the context of a funded research project. Besides PV power measurements, each time step was associated with intraday, day-ahead, and 2 days ahead forecasts by the physical model described in \cite{koster_short-term_2019}. 
We added these tier forecasts to the set of covariates used by the model ($\bold X$ in Section \ref{sec:model}) along with time covariates mentioned in Section \ref{sec:validation}, as they were meant to be available beforehand for the forecasting interval in production conditions.

The model also supports the adjunction of \emph{static} covariates, which are constant for a given sample. Relating to Section \ref{sec:model}, we note that this simply amounts to set associated $\bold x_{n,1:T}$ to a constant value. In the context of the present work, we considered a \emph{system ID} categorical feature, which is simply the system ID converted to a categorical feature with 119 modalities. We also considered \emph{system description} continuous features. Among the set of descriptors provided by system vendors and the characteristics of their setup, we retained the following features, as they were expected to influence PV power curves and magnitude: the \emph{exposition} of the system (in degrees), its \emph{inclination} (in degrees), its nominal \emph{power} (in Watts) and its \emph{calibration factor} (unitless, tied to the system on-site setup). As the DeepAR implementation expects normally distributed features, we standardized features so that they had zero mean and unit standard deviation.

As the nomimal power of our PV systems varies over a large range (from 1.4kW to 247kW), a scaling scheme was necessary to properly handle the measured values. As mentioned in Section \ref{sec:related}, we addressed this as implemented in DeepAR by dividing all measurements in a given sample by $\frac 1 {t_0} \sum_1^{t_0} {\vert z_t \vert}$. Also, as the physical model outputs were expected to be distributed similarly to their associated measurements, they were normalized in the same way.

\subsection{Loss function and metrics} \label{sec:loss}

The sum of negative log-likelihoods of observations $z_{n,t}$ (at the top of Figure \ref{fig:deepar}) was used as a loss function to fit all model parameters $\{\bold w_\mu, \bold w_\sigma, \Theta \}$ in an end-to-end fashion. 
As is common in the literature (see Section \ref{sec:ml}), we used a fixed size for the context and forecasting intervals in our experiments. As we were interested in intraday regular forecasts (see Section \ref{sec:validation}), with hourly data, the forecasting interval was size 24. 
Since it is dependent on the ECMWF NWP service, the physical model forecasts for the 3 days to come were computed each day early in the morning before sunrise, where PV power is obviously still zero. For simplicity, we thus choose midnight as the reference time of day for the regular forecasts (i.e., $t_0$ in Section \ref{sec:model}). In this context, 24-hour, 48-hour and 72-hour physical covariates associated with the predicted time step $t_0 + h$ will have been issued at time steps $t_0$, $t_0 - 24$ and $t_0 - 48$, respectively.

As the maximal horizon of the collected NWP forecasts is 72 hours, and we previously set the intraday forecasting interval as 24 hours, as a rule of thumb we used 48 hours as the context interval so that a training sample covers 72 hours. In practice, preliminary tests showed that using a larger context interval would not bring visible improvements, and using a multiple of the forecasting interval size facilitated the creation of train and test data sets.

To measure model performance, we considered metrics based on RMSE (Root-Mean-Square Error) and MAE (Mean Absolute Error).
The former are common in energy utility companies, in particular because they penalize large errors \cite{lorenz2016}, and the latter is consistent with our model, as forecasts are obtained by taking the median of sample paths \cite{yang_summarizing_2022}. We defined the normalized versions of these metrics as:

\begin{align}
    \text{nRMSE}(\hat{\bold Z}, \bold Z) & = \sqrt{\sum_{n=1}^N {\frac 1 N \frac{\frac 1 {T-t_0} \sum_{t_0+1}^T (\hat z_{nt} - z_{nt})^2}{P_n^2}}} \label{eq:rmse} \\ 
    \text{nMAE}(\hat{\bold Z}, \bold Z) & = \sum_{n=1}^N {\frac 1 N \frac{\frac 1 {T-t_0} \sum_{t_0+1}^T | \hat z_{nt} - z_{nt} | }{P_n}} \label{eq:mae}
\end{align}

with $P_n$ the nominal power of PV system $n$, $\hat {\bold Z} = \{ \hat z_{nt} \}$ the estimated point forecast, and $\bold Z = \{ z_{nt} \}$ the observed power. The normalization in Equations \eqref{eq:rmse} and \eqref{eq:mae} allowed us to measure the performance of a point estimate forecast in such a way that PV systems with larger nominal power did not dominate the error metric. This is a field requirement, as PV systems have private owners, who have to be treated equally, irrespective of the nominal power of their system. In practice, nMAE can be interpreted as a percentage of the PV system nominal power. For the sake of consistency with our model, we focused primarily on this metric, while also reporting nRMSE in Section \ref{sec:results}. To evaluate the performance of a proposed system with regards to a reference, the \emph{skill score} was derived from nMAE as:

\begin{equation}
    \text{Skill score} = 1 - \frac{\text{nMAE}(\hat{\bold Z}, \bold Z)}{\text{nMAE}(\hat{\bold Z}_\text{ref}, \bold Z)} \label{eq:skill}
\end{equation}

As stated in Section \ref{sec:model}, the models trained in the context of this work produce prediction quantiles. We used the median as the point estimate forecast; in addition, we computed the CRPS metric \cite{gneiting_calibrated_2005}, commonly used in related works \cite{gasthaus_probabilistic_2019, salinas_deepar_2020, thaker_comparative_2022}, which rates the quality of prediction quantiles as a whole:

\begin{align}
    \text{CRPS}(F^{-1}, \bold Z) &= \frac 1 {N(T-t_0)} \sum_{n=1}^N \sum_{t=t_0+1}^T \int_0^1 {2 \Lambda_\alpha (F^{-1}(\alpha), z_{nt}) d\alpha} \label{eq:crps} \\
    \Lambda_\alpha(F^{-1}(\alpha), z_{nt})& = (\alpha - \mathcal I_{[z_{nt}<F^{-1}(\alpha)]})(z_{nt} - F^{-1}(\alpha)) \nonumber
\end{align}

with $F^{-1}$ the quantile function of the predictor (which returns the quantile level in Watts associated with a probability $\alpha \in ]0,1[$), $\Lambda_\alpha$ the \emph{quantile loss}, and $\mathcal I_{[\text{c}]}$ the indicator function associated with the logical clause $c$. As discussed in Section \ref{sec:model}, the quantile function is estimated empirically using a set of sample paths $\{ \hat {\bold z}_n \}$. 
Of course CRPS penalizes observations that a long way off the median output by the model, but the penalty is increased if this difference is large while the model is excessively confident, as reflected, for example, by a narrow 90\% prediction interval (i.e., the interval between the 5\% and the 95\% quantiles). In other words, this allows excessively confident models to be penalized, and models that are able to better estimate the expected accuracy of their point forecast to be rewarded. 
In our experiments, we used 100 paths per sample as our input to return empirical quantiles. 

PV power is naturally zero at nighttime. Therefore, including these time steps in metric computation is likely to bias nMAE and CRPS towards 0. To prevent this, we excluded time steps associated with a solar zenith angle greater than $85^{o}$ \cite{lauret_benchmarking_2015} in Equations \eqref{eq:rmse}, \eqref{eq:mae} and \eqref{eq:crps}.

\subsection{Validation scheme} \label{sec:scheme}

Following recommendations by \cite{bergmeir_use_2012}, we created training samples by splitting the data set into fixed size 72-hour segments, with $t_0$ in each segment being midnight 24 hours before the end of the segment. Assuming we extracted the segment at the beginning of the available data, we then shifted the offset forward by 24 hours, so that the next segment included the previous forecasting interval in its context interval (see, e.g., top of Figure \ref{fig:rolling}). As we considered all PV systems as independent time series, this resulted in $O(CD)$ series, with $C$ the number of PV systems and $D$ the number of values $t_0$ could take in the original time series according to the segmentation method defined above.

\begin{figure}[h]
\centering
\includegraphics[width=.9\linewidth]{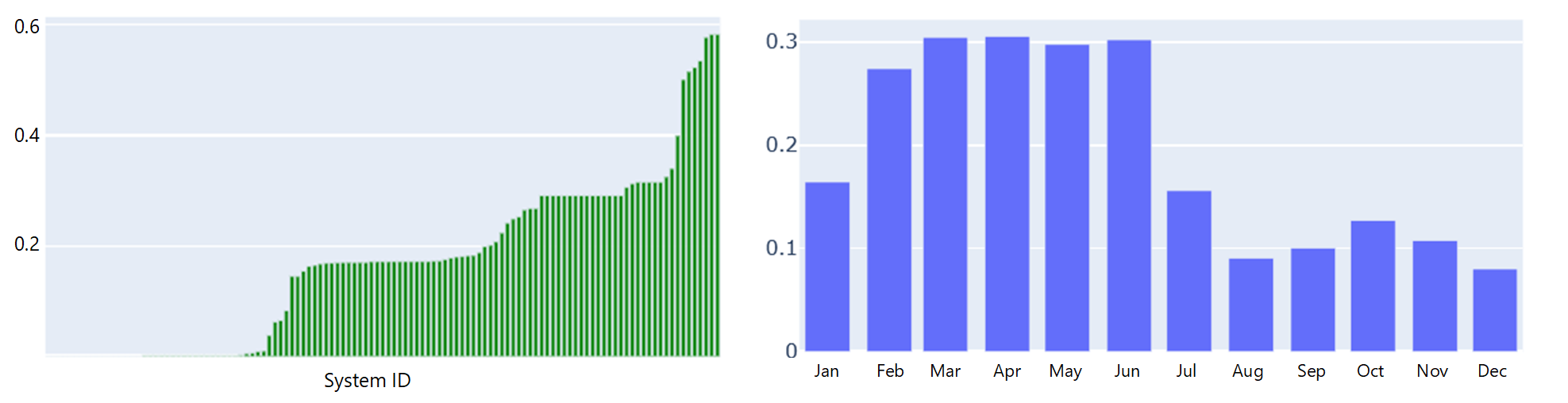}
\caption[Caption]{\emph{l.h.s.}: Proportion of missing values per system. \emph{r.h.s.}: Proportion of missing data per associated month.}
\label{fig:missing}
\end{figure}

Our number of PV systems and temporal collection bounds yielded 86,989 samples. However, PV systems may exhibit missing or erroneous measurements due to several reasons (e.g., power outage, bad manipulation, faulty sensor). Figure \ref{fig:missing} summarizes how missing values were distributed in the data set. The l.h.s. of Figure \ref{fig:missing} shows that missing values were not uniformly distributed across PV systems. Approximately one third had no missing value, another third had a bit less than 20\% of missing values, and the last third between 25\% and 50\%. The under-representation of this last third can be problematic. The r.h.s. of Figure \ref{fig:missing} shows that these missing values were not evenly distributed in time: this indicates that a group of systems may have been offline for a contiguous time frame during late winter and spring. Actually, most missing values were linked to systems that started later than the others in year 2020. The associated periods are therefore under-represented, but we note that any month has at most 30\% missing data.
In the remainder, we consider that this bias remained in a range which makes uniform sampling with regards to time acceptable for building training batches. In order to facilitate processing, and as samples cuts were aligned with day frames, we detected and excluded days matching one of the following patterns: more than two consecutive missing values, measurements blocked at a constant value, and visual inspection for aberrant values. This resulted in 67,666 valid day frames.

The PV systems are distributed in a relatively small area in Luxembourg: therefore, it was expected that forecasting intervals for different systems but with the same absolute time attached to $t_0$ would be highly correlated. In order to validate this intuition, we computed all $D$ intraday correlation matrices between systems in our data set. Specifically, we defined intraday time steps (excluding nightly time steps) as observations and PV systems as variables, resulting in $D \frac {C(C-1)} 2$ distinct correlation values. We observed that the median of the distribution of these correlation values was 0.95, which confirms a very high correlation between systems for a given day. As a consequence, naively sampling uniformly training, validation and test sets in the $O(CD)$ series would result in \emph{data leakage}, i.e., the model would be able to overfit without affecting the test error as identical samples (up to scale) would be scattered in training, validation and test sets. Let us note an unexpected positive benefit of this strong intraday correlation: the under-representation of some systems is then much less problematic. The only remaining issue pertained to estimating the parameters associated with the static categorical modalities of these systems, if using system ID static covariates. We hypothesized that at least 50\% of represented day frames is sufficient to perform this estimation.

To prevent the data leakage problem highlighted above, we first grouped the samples by the absolute time attached to their respective $t_0$, and sampled 60\% of the $D$ time steps as the training set. We used a static temporal pattern, in order to ensure that each month and season was represented fairly evenly. Validation and test sets were uniformly sampled as half of the remaining 40\%, irrespective of the PV system, which was not an issue as the goal of validation error was to be an estimate of the test error, provided parameters were not explicitly fitted on the former. The validation set was used to implement early stopping, and select the model before it started to overfit. The test set served to compute the metrics described in Section \ref{sec:loss}. To choose the cut between validation and test, and ensure that the validation error was a fair proxy of the test error, we resampled cuts until the validation and test nMAE between ground truth and intraday physical model forecasts (which are considered as constant and known in advance, and are the most relevant baseline to compare to) were equal up to a small threshold. Using the cutting procedure defined so far, we obtained 40,670 training, 13,498 validation and 13,498 test samples.

\subsection{Hyper-parameters} \label{sec:hyperparams}

\begin{figure}[h]
\centering
\includegraphics[width=1.0\linewidth]{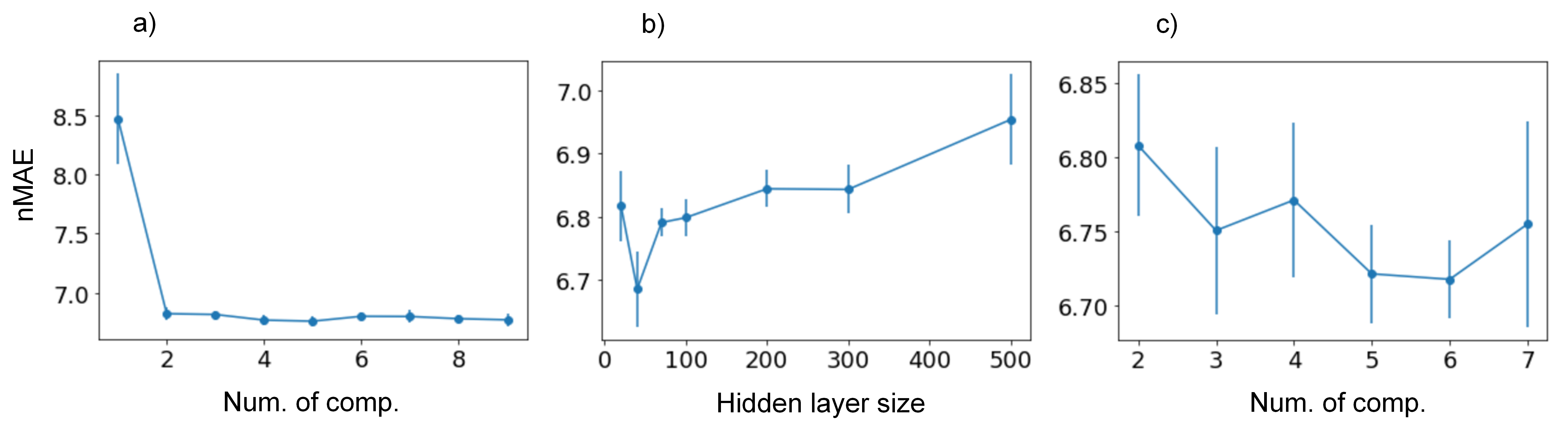}
\caption[Caption]{\emph{a)} nMAE results for variable numbers of positive Gaussian components with a hidden layer size of 100. \emph{b)} nMAE results for variable hidden layer sizes with 5 components. \emph{c)} Refinement of the number of components with a hidden layer size of 40.}
\label{fig:hyperparameters}
\end{figure}

The models were trained using the Adam optimizer with a learning rate $10^{-3}$, and a batch size of 64, for 200 epochs. Samples were reshuffled at the beginning of each epoch. In the end, we implemented a form of early stopping, by selecting the model with best validation nMAE. DeepAR uses 2 LSTM layers by default; we did not change this parametrization. Two free hyper-parameters then remained: the LSTM hidden layer size, and the number of components when a mixture distribution output is used. Figure \ref{fig:hyperparameters} shows the results of a hyper-parameter search of these parameters using the positive Gaussian distribution as mixture components, as it was the distribution used in the model that yielded the best results in Table \ref{tab:results}. Median results from 6 independently trained models are displayed in the graphs, along with $\pm 1$ standard deviation error bars. 

To avoid the brute force search of the best values of these two-hyper-parameters, we used the following heuristic to determine them jointly.
First, we first used 100 as an initial guess of the size of the hidden layers, and varied the number of components used by model 1 (see Table \ref{tab:results}, and Figure \ref{fig:hyperparameters}a). The performance was good for all but one-component models, and best for 5-component models. Then, using this number of components, we tested hidden layer sizes between 20 and 500. Results were significantly better with size 40 (Figure \ref{fig:hyperparameters}b). Finally, we refined the number of components using hidden layer size 40, and excluded one-component models as this setting was visibly irrelevant. Using 5 or 6 components yields the best results, with little difference between these two values (Figure \ref{fig:hyperparameters}c); to give a bonus to parsimony, we retained 5 components for the remainder of the experiments.

\subsection{Ablation study}

Our best model configuration (model 1 in Table \ref{tab:results}) used the positive Gaussian distribution combined with the physical model forecasts and the PV system ID as covariates. 
We first tested replacing the positive Gaussian distribution with the standard Gaussian (model 2) and Student distributions (model 3). 
We also considered continuous PV system description covariates, instead of the categorical system ID (model 4). The advantage of continuous PV system description covariates is that they enable new systems to be included without having to learn the model again, which can be highly valuable in production systems. We also tested not using PV system covariates at all (model 5), and using only one positive Gaussian component instead of a mixture (model 6). We also considered not using physical model covariates (model 7). This can be of practical interest, as access to ECMWF solar irradiance forecasts, used by the physical model as inputs, is free for research purpose, but requires a subscription for industrial applications.
Finally, we tested alternative models: MQCNN \cite{wen_multi-horizon_2018} (model 8), ISSM \cite{seeger_bayesian_2016} and FFN \cite{bebis_feed-forward_1994} (model 9). Physical and system ID covariates were used when possible (with MQCNN and ISSM, thus comparing to model 1), and were ignored otherwise (with FFN, thus rather comparing to model 7 and baselines other than the physical model).

\subsection{Results and interpretation} \label{sec:results}

Results are given in Table \ref{tab:results}. 
In \cite{koster_single-site_nodate}, skill scores were computed using a 24-hour persistence model, adjusted according to the clear sky PV power for the day under consideration. This is a common baseline in the solar energy domain \cite{pedro_assessment_2012}. For the reference, it is reported in Table \ref{tab:results}. However, in this paper, we rather considered the physical model to be the baseline against which to evaluate our results. We therefore used 24-hour physical model covariates as $\hat{\bold Z}_\text{ref}$ in Equation \eqref{eq:skill} for computing the skill scores presented in Table \ref{tab:results} (including for other baselines than the physical model). This is a stronger baseline for skill score computation, as it has been shown to significantly outperform persistence forecasts \cite{koster_short-term_2019}. In this experimental section, this means that a model with a skill score lower than 0 is not able to beat the covariates that contribute to its inputs. 

Along with the baselines mentioned above, we also report the results obtained with the Prophet model \cite{taylor_forecasting_2018} and the Non-Parametric Time Series (NPTS) forecaster \cite{gasthaus_non-parametric_2016}. Both are commonly employed as baselines in the literature. Prophet is akin to Generalized Additive Models \cite{hastie_generalized_1987}, and was fitted using an optimization algorithm. NPTS does not require any model fitting, and exploits assumptions which can be commonly done depending on time series properties (e.g., daily seasonality with hourly data). Persistence, Prophet and NPTS exploit only the input time series, and did not access the physical covariates. They could therefore be compared to models ignoring this covariate (7 and 9). 

First we can acknowledge the strength of the physical model with regards to the persistence forecast, arguably the most naive baseline. The latter has 27.1 negative skill score: this means that any model scoring less did not learn anything from the training samples better than copying the previous day as the forecast for the 24 hours to come. NPTS does very significantly better than this baseline, with skill score improved by 13.5 points. However, Prophet scores lower (-27.6), showing that this model is not suitable for the use case at hand.

\begin{table}[t]
%\scriptsize
    \centering
    \begin{tabular}{ l l l l l l }
    \textbf{ID} & \textbf{Description} & \textbf{nRMSE (\%)} & \textbf{nMAE (\%)} & \textbf{Skill (\%)} & \textbf{CRPS (-)} \\
    \hline
    \multicolumn{6}{c}{\emph{Baselines}} \\
    \hline
    \cellcolor{gray!10}- & \cellcolor{gray!10}Physical model & \cellcolor{gray!10}11.396 & \cellcolor{gray!10}7.958 & \cellcolor{gray!10}- & \cellcolor{gray!10}- \\
    - & Persistence & 16.258 & 10.117 & -27.1 & - \\
    \cellcolor{gray!10}- & \cellcolor{gray!10}NPTS \cite{gasthaus_non-parametric_2016} & \cellcolor{gray!10}$14.437(\pm 0.001)$ & \cellcolor{gray!10}$9.039(\pm 0.001)$ & \cellcolor{gray!10}-13.6 & \cellcolor{gray!10}- \\
    - & Prophet \cite{taylor_forecasting_2018} & $18.189(\pm 0.002)$ & $10.152(\pm 0.002)$ & -27.6 & - \\
    \hline
    \multicolumn{6}{c}{\emph{Models}} \\
    \hline
        \cellcolor{gray!10}1 & \cellcolor{gray!10}\textbf{Best model} & \cellcolor{gray!10}$\bold{10.808(\pm 0.082)}$ & \cellcolor{gray!10}$\bold{6.707(\pm 0.069)}$ & \cellcolor{gray!10}\textbf{15.72} & \cellcolor{gray!10}$\bold{4.816(\pm 0.048)}$ \\
        2 & w/ Gaussian & $10.950(\pm 0.031)$ & $6.844(\pm 0.070)$ & 14.00 & $4.897(\pm 0.048)$ \\ 
        \cellcolor{gray!10}3 & \cellcolor{gray!10}w/ Student & \cellcolor{gray!10}$10.928(\pm 0.082)$ & \cellcolor{gray!10}$6.841(\pm 0.047)$ & \cellcolor{gray!10}14.04 & \cellcolor{gray!10}$4.916(\pm 0.039)$ \\
        4 & wo/ system ID & $11.127(\pm 0.102)$ & $6.966(\pm 0.049)$ & 12.47 & $5.007(\pm 0.038)$ \\
        \cellcolor{gray!10}5 & \cellcolor{gray!10}wo/ system & \cellcolor{gray!10}$11.461(\pm 0.100)$ & \cellcolor{gray!10}$7.292(\pm 0.067)$ & \cellcolor{gray!10}8.37 &  \cellcolor{gray!10}$5.278(\pm 0.050)$ \\
        & wo/ mixture, &  &  &  &  \\
          \multirow{-2}{*}{6} & wo/ system & \multirow{-2}{*}{$11.547(\pm 0.042)$} & \multirow{-2}{*}{$7.711(\pm 0.017)$} & \multirow{-2}{*}{3.10} & \multirow{-2}{*}{$5.676(\pm 0.024)$} \\
        \cellcolor{gray!10}7 & \cellcolor{gray!10}wo/ physical & \cellcolor{gray!10}$14.203(\pm 0.106)$ & \cellcolor{gray!10}$9.333(\pm 0.053)$ & \cellcolor{gray!10}-17.28 & \cellcolor{gray!10}$6.577(\pm 0.032)$ \\
		8 & MQCNN \cite{wen_multi-horizon_2018} & $11.040(\pm 0.101)$ & $7.249(\pm 0.031)$ & 8.91 & $5.259(\pm 0.024)$ \\
		\cellcolor{gray!10}9 & \cellcolor{gray!10}ISSM \cite{seeger_bayesian_2016} & \cellcolor{gray!10}$12.209(\pm 0.018)$ & \cellcolor{gray!10}$8.061(\pm 0.020)$ & \cellcolor{gray!10}-1.29 & \cellcolor{gray!10}$6.231(\pm 0.002)$ \\
        & FFN \cite{bebis_feed-forward_1994} &  &  & &  \\
           & wo/ system, &  &  &  &  \\
          \multirow{-3}{*}{10} & wo/ physical & \multirow{-3}{*}{$14.687(\pm 0.069)$} & \multirow{-3}{*}{$10.087(\pm 0.033)$} & \multirow{-3}{*}{-26.75} & \multirow{-3}{*}{$7.522(\pm 0.037)$} \\
    \end{tabular}
    \caption{nRMSE, nMAE and CRPS test metrics for the range of compared models. The results of the best performing model are bold-faced. Median results and standard deviations were estimated from 6 models trained independently. \emph{w/} and \emph{wo/} stand for \emph{with} and \emph{without}, respectively.}
    \label{tab:results}
\end{table}

% noting that variance is much lower with proposed solution
% Raw FFN not much better than LSTM, but the latter allows to use covariates effectively
% but significant CRPS degradation

Then, attending the ablation study results, we see that using the positive Gaussian component has a small but significant influence, with skill scores improved by 1.72 and 1.68 points with respect to the Gaussian and Student components, respectively. The performance of model 4 using the continuous PV system description features also yielded solid performance, with a 2.95 points score degradation compared to model 1. 
We note that the relatively better performance observed by using the system ID as a covariate provides anecdotal evidence supporting the hypothesis formulated in Section \ref{sec:scheme} regarding the imbalance of systems representation in the dataset.
The first significant performance gap was seen with model 5, which ignored all PV system covariates. Its score was 4.10 points lower than model 4. Then, as could be anticipated from Section \ref{sec:hyperparams} on hyper-parameters optimization, using a single positive Gaussian component degraded the performance of model 5 by 5.27 additional points. Finally, the largest gap was observed when ignoring the physical model covariates, but otherwise keeping the same configuration as model 1, with a negative skill score of -17.28\%. This shows that the neural model alone is not able to compete with the physical model, and implicitly validates the relevance of the \emph{hybrid-physical} approach, which aims to build upon physical models with machine learning techniques \cite{antonanzas2016}. We also note that its skill score is better than persistence, but worst than NPTS (-17.28 vs. 13.6 for the latter). This shows that strong non-parametric baselines are always worth attending.
Ultimately, using MQCNN yielded a positive skill score, but degraded by 6.81 points with respect to model 1. Despite accessing physical covariates and PV system IDs, ISSM obtained a negative skill score, showing its low relevance for the type of data we considered in this work. Finally, using FFN degraded the results of model 7, its closest variant in our ablation study in terms of available input data, by 9.47 points. This validated our design based on DeepAR and our positive Gaussian component.

\begin{figure}[h]
\centering
\includegraphics[width=1.0\linewidth]{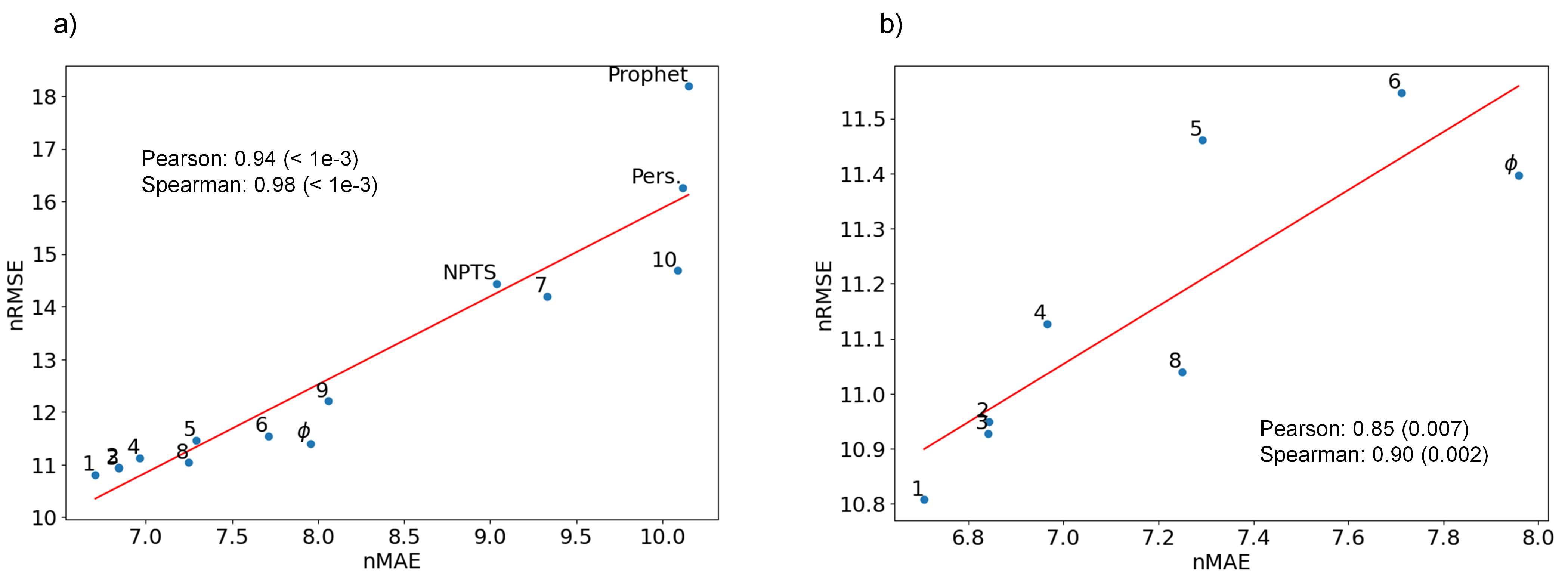}
\caption[Caption]{Illustration of the relationship between nMAE and nRMSE metrics \emph{(a)}, also magnified after excluding models with nMAE greater than 9\% from the representation \emph{(b)}. $\phi$ stands for the baseline physical model. Red lines are the result of a linear regression of the points in the graph, reported along with Pearson and Spearman correlation coefficients.}
\label{fig:mae-vs-rmse}
\end{figure}

In Figure \ref{fig:mae-vs-rmse}a, we see that all models not using physical model covariates have the largest nMAE, and thus have an important leverage on any analysis of the correlation between nMAE and nRMSE metrics. Indeed, excluding them had a significant effect on the regression line and computed coefficients (Figure \ref{fig:mae-vs-rmse}b), even if the correlation remained high and significant. Models below the regression line tended to have a lower nRMSE than would be expected according to their nMAE. This was the case for all models using the system ID (1, 2, 3 and 8), as well as the physical model baseline. This means that in some production contexts that would favor nRMSE as a prime metric, using the system ID as a covariate is preferable. As a side note, we also computed the analysis shown in Figure \ref{fig:mae-vs-rmse} between nMAE and CRPS metrics; all correlation coefficients were then highly significant and beyond 0.96, even when excluding models with nMAE above 9\%. This means that in our experiments, nMAE and CRPS were almost perfectly correlated. Reporting specifically on CRPS would therefore not bring significant added value.

As alternative designs, we considered using unscaled physical model covariates (i.e. not scaling them along with $\bold z$ values as described in Section \ref{sec:data}), and using the weighted sampling scheme described in \cite{salinas_deepar_2020}, which samples training examples with a frequency proportional to the magnitude of $\bold z$ values. We did not report the results of these alternative designs as they brought systematic degradation to the performance. We also tried to use both types of system description covariates (i.e., system ID and continuous description features) simultaneously, but this led to a slight degradation compared to the respective model using only the system ID covariate. This is expected, as the system ID alone already encodes system diversity. In addition, as we will see in the next section, continuous system description features may reflect the system setup in a biased way, which could explain the slight degradation in performance observed in Table \ref{tab:results}.

\section{Discussion and qualitative examples} \label{sec:discussion}

We first recall that the contributions highlighted in the introduction are distributed throughout the paper, not only in terms of our theoretical contributions in Section \ref{sec:model}, which are arguably modest, as we position our work as an application, but also in our analysis of the related work, notably positioning hybrid-physical approaches as a special case in the context of neural-network-based time series forecasting in Section \ref{sec:ml}. 
We also highlight practical contributions to data preprocessing, training and validation procedures in Sections \ref{sec:ml} and \ref{sec:scheme} which will enable the transfer of the work to the DSO.

In the previous section, we evaluated our proposed models using global metrics. In this section, we aim to provide a more detailed insight into our results by analyzing model performance at the system level. The examples displayed were obtained using the best performing model (i.e., model 1 in Table \ref{tab:results}). First, we computed nMAE metrics for each sample in the test set, grouped them according to their associated system ID, and ranked the systems according to the difference between model 1 and the physical model nMAE. In other words, the higher a system in this ranking, the better model 1 outperforms the physical model for this system. For all but 1 system of 119, model 1 performs better than the baseline. We first consider the worst case, as seen in system 68.

\begin{figure}[h]
\centering
\includegraphics[width=.9\linewidth]{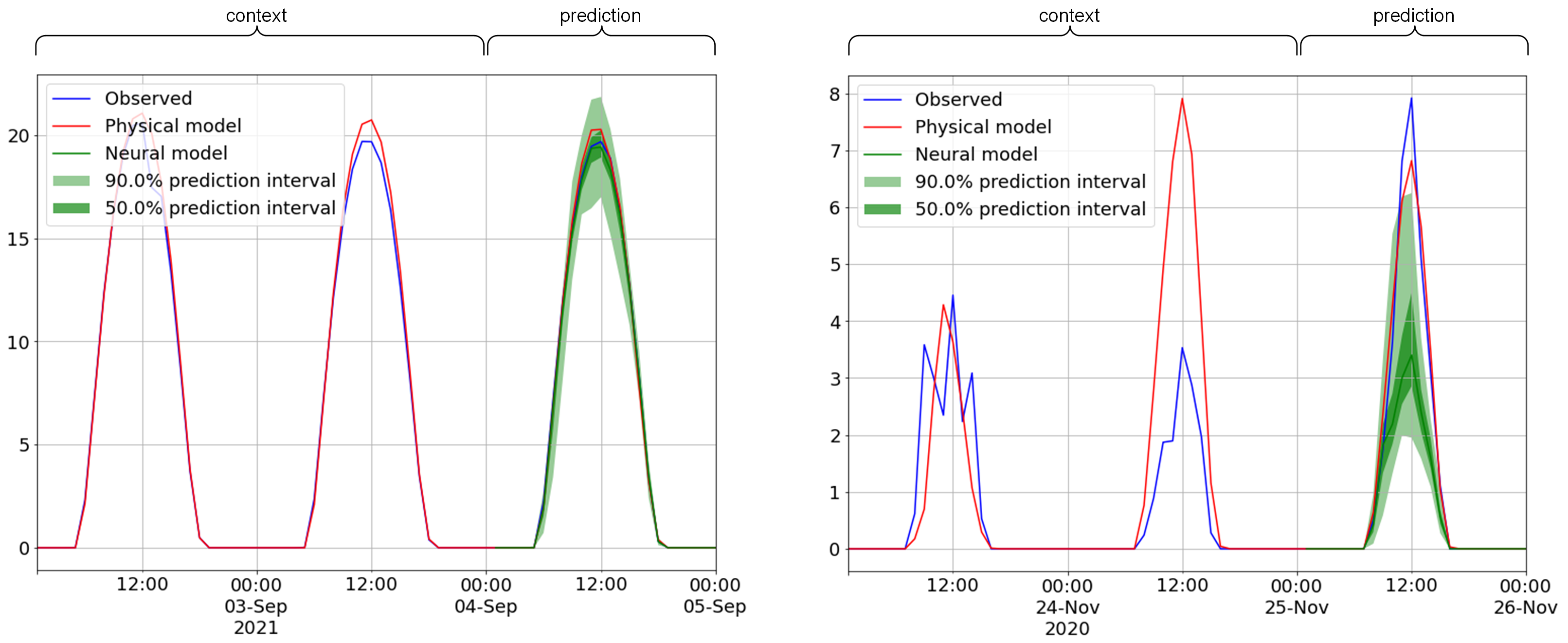}
\caption[Caption]{Two test samples associated with system 68. Prediction intervals are displayed using green shades. The observations and 24-hour physical model forecasts of the context interval are prepended.}
\label{fig:worst-case}
\end{figure}

\begin{figure}[h]
\centering
\includegraphics[width=.9\linewidth]{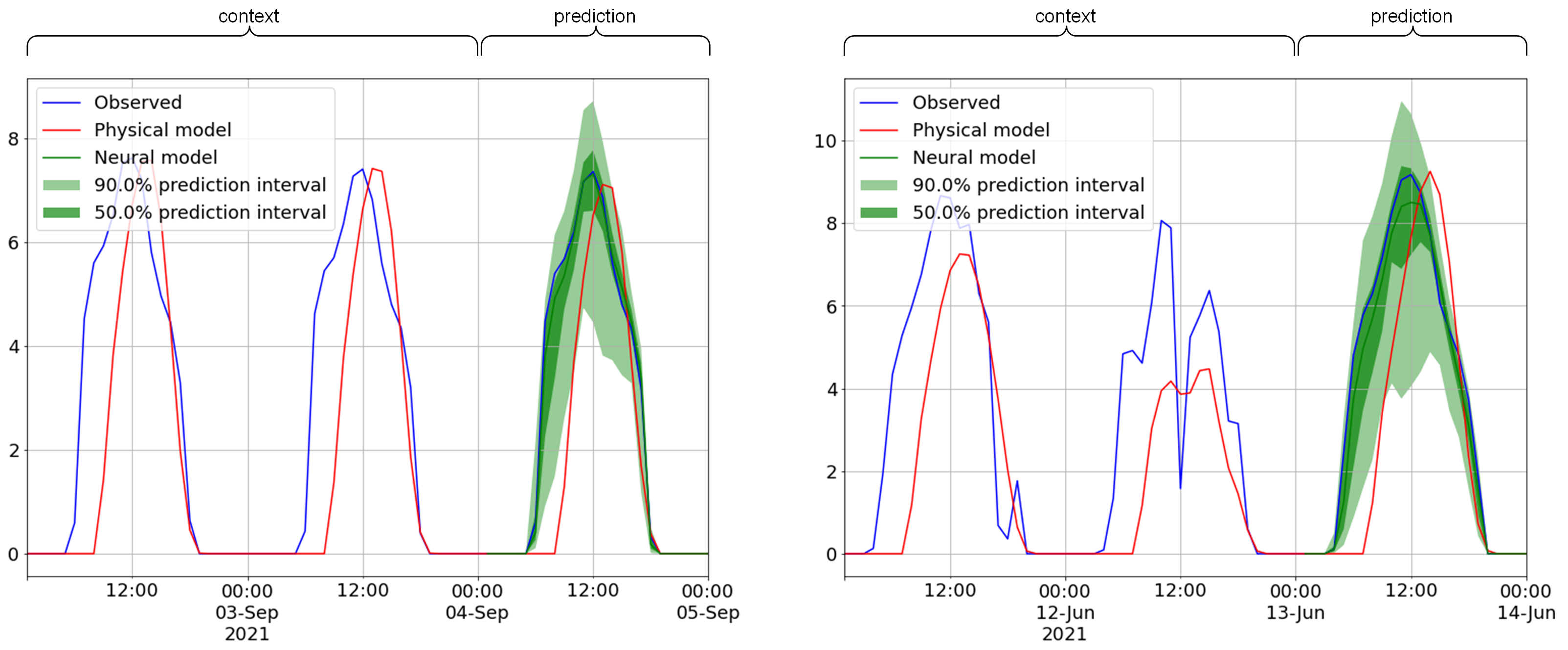}
\caption[Caption]{Two test samples associated with system 44. Prediction intervals are displayed using green shades. The observations and 24-hour physical model forecasts of the context interval are prepended.}
\label{fig:best-case}
\end{figure}

On the l.h.s. of Figure \ref{fig:worst-case}, we display the sample of this system that has among the lowest nMAE with the DeepAR model. This is a typical clear day example, where the prediction is fairly easy for the neural model. We note that for this instance, forecasts stick more closely to the observation curve than the baseline. The prediction intervals are naturally tight, reflecting the high confidence of the neural model in this case. On the r.h.s. of Figure \ref{fig:worst-case}, for the same system, we display a sample with one of the biggest differences between the two models. In this case, DeepAR is not able to keep up with the sudden peak of PV power. While the 24-hour physical model covariate provided information on this peak, this information was not used by model 1, which acted conservatively regarding the observations in the context interval.

In Figure \ref{fig:best-case}, we consider samples from system 44. This system is the third best in our ranking, and has been identified as problematic for the physical model because of a double-pitched roof, which is not reflected in the system description features \cite{koster_single-site_nodate}. On both sides of Figure \ref{fig:best-case}, the systematic shift of the PV performance model is clearly visible. We also see that model 1 is able to completely ignore this shift, and still come up with sensible forecasts. The figure also shows how the prediction intervals are narrower when the PV production is stable over several days (l.h.s.), and wider when the weather seems more unstable (r.h.s.).

\section{Conclusion}

Ultimately, we were able to improve power forecasts obtained from an already strong physical PV performance model. Our experiments, comprising an ablation study and model comparisons, highlight the best working configuration, which uses the PV performance model forecasts as covariates, a mixture of positive Gaussians as the output distribution, and a static categorical covariate that reflects the associated system ID. The positive Gaussian output is effective for the \emph{bell-shaped} data profile typical of solar energy applications, and the system ID feature allows local effects, which went previously unnoticed with the physical PV performance model alone, to be modeled.

In future works, we plan to refine and explore novel neural model designs. For example, quantile regression methods more recent than \cite{wen_multi-horizon_2018} will be explored. Also, we will further investigate how to add novel systems to the grid without having to retrain the full model. We saw that using system description features is an effective fallback, but these features do not account for local effects such as a double-pitched roof and therefore remain suboptimal. We will also consider longer forecasting intervals (e.g. day-ahead and 2 days ahead).

In our experiments, early stopping was critical to obtain models with good generalization abilities, as without monitoring the validation error after each epoch, our training procedure tended to overfit. This is mostly due to the measures we took to prevent data leakage: when segmenting 2 years of data at the day scale, despite all our efforts, training and test sets are unlikely to be identically distributed. We addressed this problem in the most straightforward and conservative way, but it seems to be related to the domain shift problem characterized by the domain adaptation literature \cite{redko2019}. Adapting contributions from this area to the distinctive features of our application is a task for future works.

\subsection{Acknowledgements}

This research was funded by the Luxembourg National Research Fund (FNR) in the framework of the FNR BRIDGES \emph{CombiCast} project with grant number \newline (BRIDGES18/IS/12705349/Combi-Cast). Furthermore, the authors would like to thank our partner Electris (a brand of Hoffmann Frères Energie et Bois s.à r.l.), for its trust, the very supportive partnership throughout the whole project duration, and its contribution to the common project, both financially and in terms of manpower and data.

\subsection{Data Availability}

The datasets analyzed during the current study are not publicly available as they belong to Electris. A sample may be provided by contacting the corresponding author with a reasonable request, without any guarantees.

\subsection{Declarations}

\subsubsection{Conflicts of interest} 

All authors declare that they have no conflicts of interest.

\bibliography{refs}

\end{document}